\documentclass[conference]{IEEEtran}
\IEEEoverridecommandlockouts
\usepackage{cite}
\usepackage{amsmath,amssymb,amsfonts}
\usepackage{algorithm}
\usepackage{algorithmic}
\usepackage{graphicx}
\usepackage{textcomp}
\usepackage{xcolor}
\usepackage{xspace}
\usepackage{soul}
\usepackage{multirow}
\usepackage{booktabs}
\usepackage{caption}
\newcommand{\BfPara}[1]{\vspace{1mm}{\noindent\bf#1.}\xspace}

\def\BibTeX{{\rm B\kern-.05em{\sc i\kern-.025em b}\kern-.08em
    T\kern-.1667em\lower.7ex\hbox{E}\kern-.125emX}}
\begin{document}

\title{Trends in Neural Architecture Search: \\ Towards the Acceleration of Search}

\author{\IEEEauthorblockN{$^{\dag}$Youngkee Kim, $^{\dag}$Won Joon Yun, $^{\circ}$Youn Kyu Lee, $^{\ddag}$Soyi Jung, and $^{\dag}$Joongheon Kim}
\IEEEauthorblockA{
$^{\dag}$Department of Electrical and Computer Engineering, Korea University, Seoul, Republic of Korea 
\\
$^{\circ}$Department of Computer Engineering, Hongik University, Seoul, Republic of Korea
\\
$^{\dag}$School of Software, Hallym University, Chuncheon, Republic of Korea 
\\
E-mails: 
\texttt{felixkim@korea.ac.kr}, 
\texttt{ywjoon95@korea.ac.kr}, 
\texttt{younkyul@hongik.ac.kr}, \\
\texttt{jungsoyi@korea.ac.kr}, 
\texttt{joongheon@korea.ac.kr}
}
}
\maketitle

\begin{abstract}
In modern deep learning research, finding optimal (or near optimal) neural network models is one of major research directions and it is widely studied in many applications.
In this paper, the main research trends of neural architecture search (NAS) are classified as neuro-evolutionary algorithms, reinforcement learning based algorithms, and one-shot architecture search approaches. Furthermore, each research trend is introduced and finally all the major three trends are compared. Lastly, the future research directions of NAS research trends are discussed.
\end{abstract}

\section{Introduction}

Recent years, deep learning methods are actively and widely used for many applications such as object detection, image recognition, and natural language processing~\cite{pieee202105park}.
As observed in many deep learning methods, they are fundamentally for heuristic-based nonlinear function approximation, therefore they cannot guarantee optimal solutions~\cite{tvt201905shin}. 
Hence, finding the optimal deep learning models is regarded as a hyper-parameter tuning problem depending on given tasks. 
Conventionally, skilled experts have resolved the problems through costly trial-and-error processes based on their experiences. This repetitive process involves the design of neural network architectures that accounts for most of overall costs. Thus, the cost for designing a new deep learning model can be reduced due to the minimization of manual work of human experts.
In this perspective, \emph{Neural Architecture Search (NAS)} was emerged in modern deep learning research. The objective of NAS is to obtain high-performance neural architectures at low search costs thanks to the automation of the neural architecture design process. 
One of the first research results that introduced the concept of NAS was proposed by Zoph \textit{et. al.}~\cite{zoph2016neural}. A reinforcement learning based approach is applied for NAS, and the neural architectures derived from the algorithm in~\cite{zoph2016neural} have achieved the state-of-the-art performance human designed. Since then, many corresponding and related research contributions have been discussed and actively developed. 

In this paper, we introduce and compare the research trends in NAS; and finally discuss future research directions.
We first look at the trends of NAS research activities in Sec.~\ref{sec:sec2}, then classify and summarize major approaches in Sec.~\ref{sec:sec3}. After that, Sec.~\ref{sec:sec4} discusses the corresponding future research directions and compares the existing algorithms. Sec.~\ref{sec:sec5} concludes this paper.

\section{Motivation and Problem Statements}\label{sec:sec2}

In early NAS research results, one of the most important performance indicators was the accuracy of the final model derived from each search method. So far, the effectiveness of NAS has been proven by a number of experiments that achieved better performance than human designed architectures. Nevertheless, it is not easy to apply NAS techniques to various fields due to resource and time constraints. To address this limitation, recent studies focus on the reduction of search costs while preserving the performance of deliverables. In this trend, we introduce major approaches of NAS.

\begin{figure}[t!]
    \includegraphics[width=1\columnwidth]{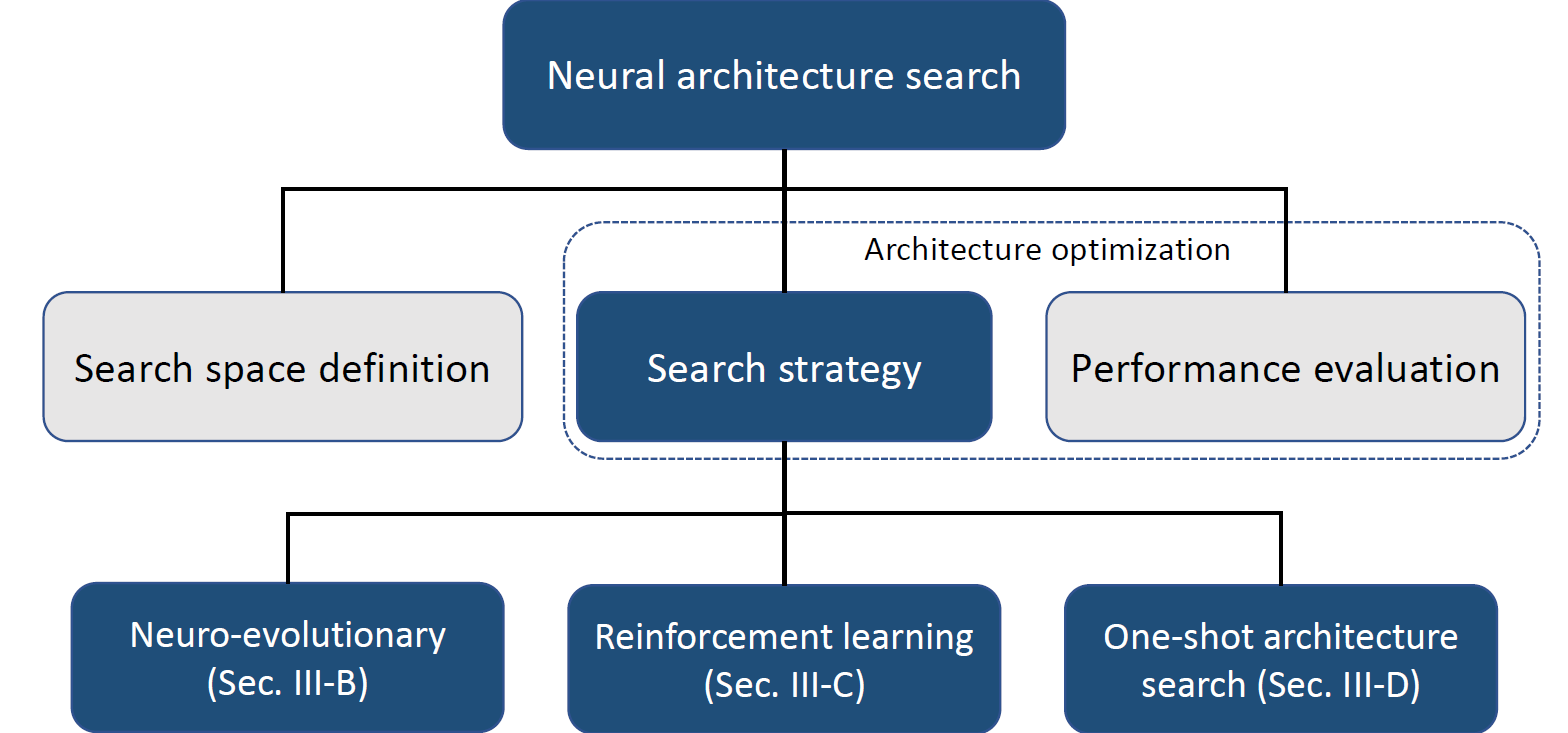}
    \caption{Taxonomy of NAS research areas.}
    \label{fig:taxonomy}
\end{figure}

\section{NAS: Towards the Acceleration of Search}\label{sec:sec3}

\subsection{Taxonomy}

In this paper, the research topics of NAS are organized in the way illustrated in Fig.~\ref{fig:taxonomy}. From a macroscopic perspective, NAS components can be categorized as the search space definition, search strategy, and performance evaluation criterion. The definition of search space can be considered as the specification of scope of possible candidate neural architectures. This means that the wider search space is required as more neural architectures are taken into account. Eventually, it leads to considerable search costs. Several studies have been conducted to address this trade-off, however these are out of scope in this paper.
We focus on the approaches in terms of neural architecture optimization that includes search strategy and performance evaluation criterion, but the following section more focuses on the search strategy.

\subsection{Approach 1: Neuro-evolutionary algorithms}
\begin{figure}[t!]
    \includegraphics[width=1\columnwidth]{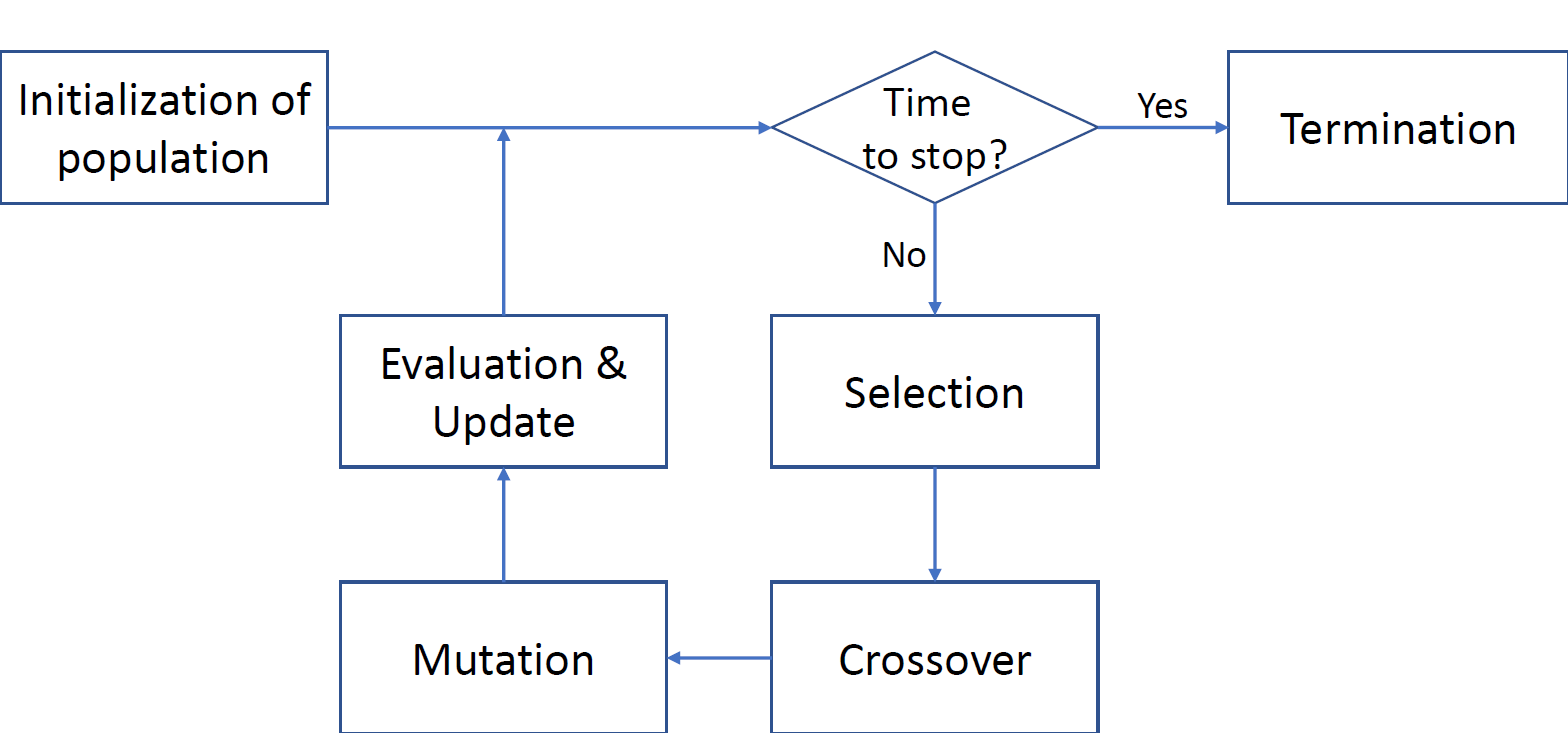}
    \caption{Flowchart of evolutionary algorithm.}
    \label{fig:evolutionary}
\end{figure}

The approach of search strategy for NAS is a neuro-evolutionary algorithm. The neuro-evolution is a method that utilizes evolutionary algorithms to generate neural networks. For the first time, Miller \textit{et. al.}~\cite{miller1989designing} proposed a method that designs neural network architectures using genetic algorithms and also optimizes the parameters of networks with back-propagation. Many follow-up studies apply this neuro-evolutionary approach in order to discover a novel neural network architecture. Fig.~\ref{fig:evolutionary} illustrates the flowchart of general evolutionary algorithm computation procedure. The algorithm conducts iterative computation until satisfying predefined conditions such as search time thresholds and target performances. 
It starts with the initial population composed of simple neural network models. In every single evolution step, parent models are sampled from the population (\textit{Selection} phase) and a new architecture is derived that inheriting the information of each parent model (\textit{Crossover} phase). The offspring architectures are mutated in various ways such as adding or removing some layers and changing architectural parameters (\textit{Mutation} phase). This newly created architecture is determined whether to be thrown away or registered with the population according to the performance after training (\textit{Evaluation \& Update} phase). The parent model sampling methods, mutation techniques and population updates are major research topics for neuro-evolutionary algorithms.

More recently, Real \textit{et. al.}~\cite{real2019regularized} introduced a neural network for image classification tasks called \emph{AmoebaNet-A}, which is obtained through architecture search using an improved evolutionary algorithm. They applied the age property to the tournament selection phase of the algorithm to give a tendency to favor a new generation. This improvement leads the AmoebaNet-A to achieve a new state-of-the-art top-5 ImageNet accuracy. 

\subsection{Approach 2: Reinforcement learning (RL)-based algorithms}
\begin{figure}[t!]
    \includegraphics[width=1\columnwidth]{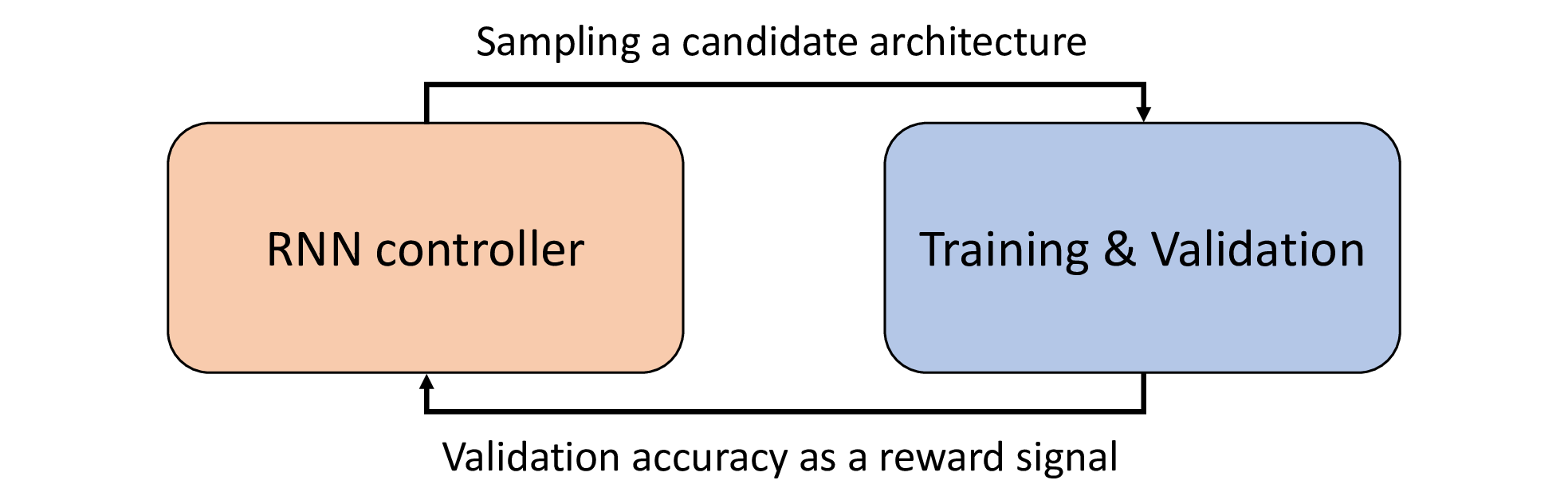}
    \caption{Overview of RL-based NAS.}
    \label{fig:rl}
\end{figure}

Another approach for exploring the search space is fundamentally based on reinforcement learning (RL). This approach addresses the optimization for architecture search problem from an RL perspective. Baker \textit{et. al.}~\cite{baker2016designing} conducted a study to generate high-performance convolutional neural network (CNN) architectures with \(\varepsilon\)-greedy Q-learning algorithm. NAS-RL~\cite{zoph2016neural} is one of the first attempts to apply RL-based approaches to NAS. In NAS-RL, neural network architectures are represented by a type of variable-length string. This representation allows a recurrent neural network (RNN) based controller to sample a candidate architecture by the sequential inference for architectural parameters. The validation accuracy of sampled neural architecture is obtained though training and validation processes. This validation accuracy serves as a reward signal for training the controller by policy gradient methods. The overall learning process for the controller is depicted in Fig.~\ref{fig:rl}.

As a subsequent study, Zoph \textit{et. al.}~\cite{zoph2018learning} introduced a method to generate a neural network for image classification, called NASNet, which is based on the RL search strategy. While the proposed method is similar to NAS-RL, the proximal policy optimization (PPO) is applied to the controller training. Furthermore, they adopted a cell-based search space and \emph{ScheduledDropPath}, a novel regularization technique, which enables the transferability and improves the generalization in NASNet. Although NASNet is only trained on CIFAR-10, it achieves reasonable accuracy on ImageNet with a slight tuning. 

\subsection{Approach 3: One-shot architecture search}
With one-shot architecture search approaches, we can try alternative approaches rather than conventional schemes. In early NAS works, the training process for each candidate architecture is typically high-cost. Based on this observation, the key objective of one-shot architecture search is to minimize the training costs during the search. 
The concept of one-shot architecture search is so comprehensive, that we describe some representative methods in this paper.

\begin{figure}[t!]
    \includegraphics[width=1\columnwidth]{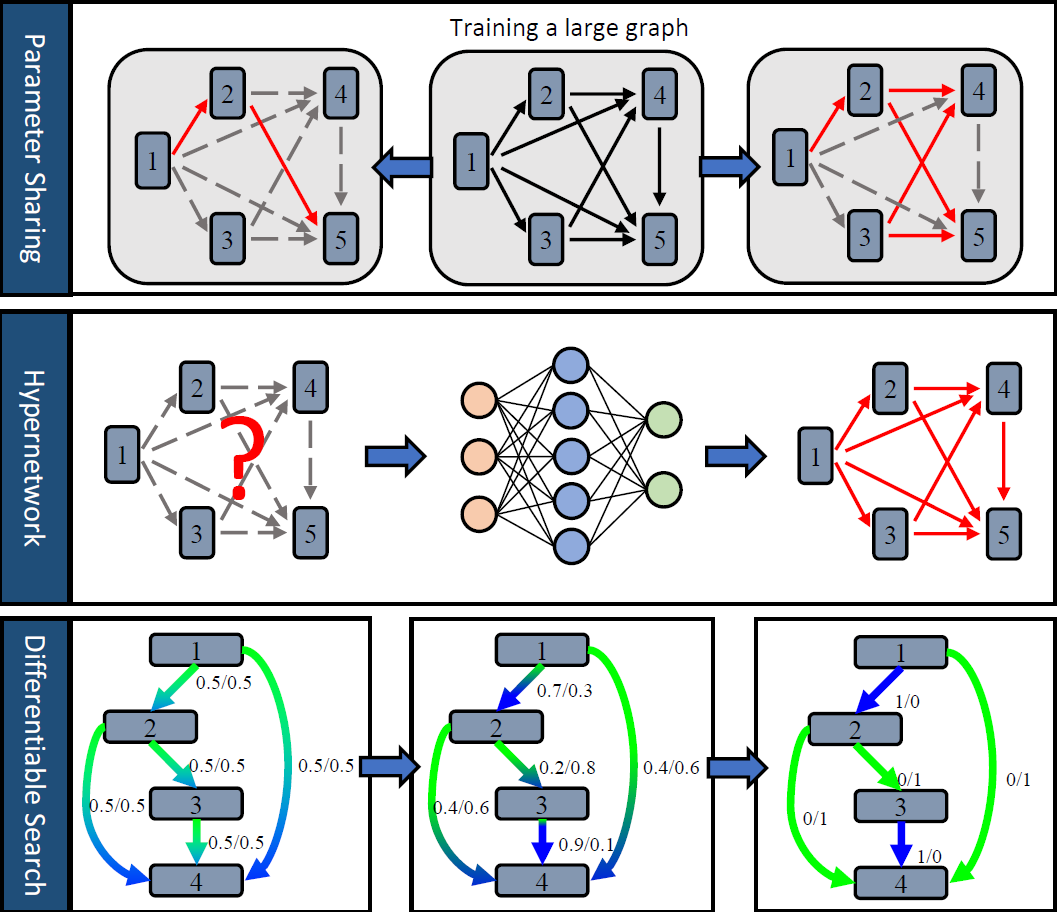}
    \caption{One-shot architecture search approaches.}
    \label{fig:oneshot}
\end{figure}

\BfPara{Parameter Sharing} The parameter sharing is one of the most widely used methods in deep learning. From the view point of NAS, it is clear that training each candidate architecture from scratch results in huge computing costs. In other words, it is a great waste to throw away the knowledge acquired in each training of the candidates. First of all, ENAS~\cite{ENAS} is one of the initial studies to bring the parameter sharing to NAS. The entire search space of ENAS is regarded as an over-parameterized network, which is represented by a directed acyclic graph. An LSTM-based controller searches on the subgraphs of this large graph to obtain the optimal neural network architecture. This approach allows all candidates to share their parameters eliminating the need for training from scratch. The first row in Fig.~\ref{fig:oneshot} depicts the subgraph sampling with parameter sharing.

\begin{table*}[t!]
\centering
\resizebox{\textwidth}{!}{\begin{tabular}{@{}ccclccc@{}}
\toprule
\multicolumn{1}{c}{Approach} & \multicolumn{1}{c}{Advantage} & \multicolumn{1}{c}{Disadvantage} & \multicolumn{1}{c}{Reference} & \multicolumn{1}{c}{Accuracy} & \multicolumn{1}{c}{Parameters} & \multicolumn{1}{c}{GPU days} \\ \midrule
\multirow{2}{*}{\begin{tabular}[c]{@{}c@{}}Neuro-evolutionary Algorithm\end{tabular}} & \multirow{2}{*}{Robustness} & \multirow{2}{*}{Slow search time} & \multirow{2}{*}{AmoebaNet-A\cite{real2019regularized}} & \multirow{2}{*}{96.66 ${\pm}$ 0.06\%} & \multirow{2}{*}{3.2M} & \multirow{2}{*}{3150} \\
 &  &  &  &  &  &  \\ \midrule
\multirow{2}{*}{Reinforcement Learning} & \multirow{2}{*}{Guaranteed performance} & \multirow{2}{*}{Slow search time} & \multirow{2}{*}{NASNet-A + cutout\cite{zoph2018learning}} & \multirow{2}{*}{97.35\%} & \multirow{2}{*}{3.3M} & \multirow{2}{*}{2000} \\ 
 &  &  &  &  &  &  \\ \midrule
\multirow{3}{*}{One-shot Architecture Search} & \multirow{3}{*}{Fast search time} & \multirow{3}{*}{Huge memory space} & ENAS + cutout\cite{ENAS} & 97.11\% & 4.6M & 0.45 \\ \cmidrule(l){4-7} 
 &  &  & GHN\cite{zhang2018graph} & 97.16 ${\pm}$ 0.07\% & 5.7M & 0.84 \\ \cmidrule(l){4-7}
 &  &  & DARTS(first order)+cutout\cite{DARTS} & 97.00 ${\pm}$ 0.14\% & 3.3M & 1.5 \\\bottomrule
\end{tabular}}
\caption{Summary of major approaches in NAS.}
\label{table:tbl}
\end{table*}

\BfPara{Hypernetwork} The hypernetwork was proposed in \cite{ha2016hypernetworks} as a neural network to generate non-shared weights for other larger networks such as CNN and LSTM. Brock \textit{et. al.}~\cite{brock2017smash} applied this technique to predict the performance of candidate architectures. The middle row in Fig.~\ref{fig:oneshot} visualizes the role of a hypernetwork that learned to generate the weights for a given architecture. This hypernetwork can rank candidate architectures and determine the best of them. More recently, Zhang \textit{et. al.}~\cite{zhang2018graph} improved this approach with the concept of a graph neural network. They proposed the Graph HyperNetwork (GHN) that predicts the performance more accurately by modeling the topology of a given architecture.

\BfPara{Differentiable search} Most NAS approaches have the underlying premise that the optimization problem for architecture search is a black-box optimization over a discrete domain. This fact naturally results in a large amount of computation and search time as the search space grows. To address this problem, DARTS~\cite{DARTS} emerged with a different perspective on the definition of search space. DARTS, as its name implies, have a differentiable search space by applying the continuous relaxation to the architecture representation. The process of obtaining the optimized cell unit structure is depicted in the bottom of Fig.~\ref{fig:oneshot}. The cell is represented by a directed acyclic graph that each node means a latent representation and each edge indicates a mixture of candidate operations having continuous probability. These parameterized probability and the weights of entire architecture are optimized simultaneously by resolving the bilevel optimization problem with gradient descent method.

\section{Discussions and Future Research Directions}\label{sec:sec4}


In this section, we will discuss which approach is better starting point for future research in terms of search time by summarizing the pros and cons of each approach.

The first approach that we covered was neuro-evolutionary algorithms. The benefits of this approach are inherently similar to the evolution. It allows to search a neural architecture for a high robustness regardless the nature of problems. In addition, this population-based method is able to explore multiple parts of the search space in parallel, offering opportunities to avoid the local optima. On the other hand, the neuro-evolutionary algorithm has weaknesses in computational cost and learning speed. The number of evaluation steps increases drastically by the size of population and training samples. In addition, the learning through generations takes considerable amount of times. 

The next is reinforcement learning based approaches that train the agent to generate well-performing neural architectures. Learning principles of RL-based algorithms are similar to those of human being. Thus, it is relatively clear that the final model outperforms that designed by the human expert. However, a significant search time is inevitable due to repetitive action and reward processes.

As explained in previous section, The goal of one-shot architecture search is to minimize the training cost of candidate neural architectures. It is easy to predict that this motivation leads to save the total search time. As shown in Table~\ref{table:tbl}, the one-shot architecture search algorithms achieved significant improvements in terms of search time. The performance metrics in this table are based on CIFAR-10. However, this approach implies the training of a large computational graph or the optimization for the entire search space at once. Consequently, the one-shot architecture search methods require greater memory space than other approaches.


Although accelerating the search process is still one of major improvement points, it is time to consider how to apply these accelerated NAS technologies to various domains in a practical way. NAS-Unet~\cite{weng2019unet} is one of studies that extends NAS into medical domain. As its name suggests, NAS-Unet is inspired by U-Net~\cite{unet}, which is widely used for image segmentation in biomedical. The NAS-Unet is discovered by the proposed NAS method that applies a differentiable architecture search strategy based on a U-like backbone architecture. It attains better performance and much fewer parameters than U-Net without any pretraining.


\section{Concluding Remarks}\label{sec:sec5}

In this paper, the main streams of NAS stuides are classified as neuro-evolutionary algorithms, reinforcement learning based algorithms, and one-shot architecture search approaches. Furthermore, each research direction is summarized as a core idea, along with representative study cases. With this summary, researchers interested in NAS can obtain base domain knowledge to set a good starting point.

\section*{Acknowledgment}
This work was supported by Institute of Information \& Communications Technology Planning \& Evaluation (IITP) grant funded by the Korea government(MSIT) (No. 2021-0-00766, Development of Integrated Development Framework that supports Automatic Neural Network Generation and Deployment optimized for Runtime Environment). Joongheon Kim is a corresponding author of this paper.

\bibliographystyle{IEEEtran}
\bibliography{ref_nas,ref_aimlab}
\end{document}